\documentclass[letterpaper, 10 pt, conference]{ieeeconf}  

\IEEEoverridecommandlockouts                              

\usepackage{amsmath} 
\usepackage{graphicx}
\usepackage{multirow}
\usepackage{url}
\usepackage{cite}

\title{\LARGE \bf
Learning Quadrupedal Locomotion for a Heavy Hydraulic Robot Using an Actuator Model
}

\author{Minho Lee$^{1}$, Hyeonseok Kim$^{2}$, Jin Tak Kim$^{3}$, Sangshin Park$^{3}$, Jeong Hyun Lee$^{1}$, Jungsan Cho$^{3}$, \\and Jemin Hwangbo$^{*, 1}$
\thanks{This work was supported by the Korea Research Institute for defense Technology planning and advancement(KRIT) grant funded by the Korean government(DAPA(Defense Acquisition Program Administration)). (No. 20-107-C00-007-02(KRIT-CT-22-001), Development of Walking-Driving Hybrid Locomotion Control for Multi-Legged Robot Platform, 2024)}
\thanks{$*$ corresponding author}
\thanks{$^{1}$ Robotics and Artificial Intelligence Lab, KAIST, Daejeon, South Korea}%
\thanks{$^{2}$ Robotics Team, HYUNDAI Rotem, Uiwang, Gyeonggi-do, South Korea}%
\thanks{$^{3}$ Hydraulic Robot Laboratory, Human-Centric Robotics R\&D Department, Korea Institue of Industrial Technology, Ansan, Gyeonggi-do, South Korea}%
\thanks{{\tt\small myno1126@kaist.ac.kr, }}
\thanks{{\tt\small hyunseok0427@hyundai-rotem.co.kr, jintagi@kitech.re.kr, pss@kitech.re.kr, josualee@kaist.ac.kr, chojs@kitech.re.kr, jhwangbo@kaist.ac.kr}}
}

\usepackage{hyperref}

\begin{document}

\maketitle
\thispagestyle{empty}
\pagestyle{empty}

\begin{abstract}

The simulation-to-reality (sim-to-real) transfer of large-scale hydraulic robots presents a significant challenge in robotics because of the inherent slow control response and complex fluid dynamics. The complex dynamics result from the multiple interconnected cylinder structure and the difference in fluid rates of the cylinders. These characteristics complicate detailed simulation for all joints, making it unsuitable for reinforcement learning (RL) applications. In this work, we propose an analytical actuator model driven by hydraulic dynamics to represent the complicated actuators. The model predicts joint torques for all 12 actuators in under 1 microsecond, allowing rapid processing in RL environments. We compare our model with neural network-based actuator models and demonstrate the advantages of our model in data-limited scenarios. The locomotion policy trained in RL with our model is deployed on a hydraulic quadruped robot, which is over 300 kg. This work is the first demonstration of a successful transfer of stable and robust command-tracking locomotion with RL on a heavy hydraulic quadruped robot, demonstrating advanced sim-to-real transferability.

\end{abstract}


\section{INTRODUCTION}

Quadruped robots have been widely used for navigating through various obstacles and complex terrains \cite{winkler2015terrain, rong2012HydQuad, jeon2024wholebody, minihyq, kuindersma2016atlas, lee2024traversability, kim2025parkour}. They have demonstrated stability \cite{minihyq} and terrain adaptability \cite{winkler2015terrain, leeJoonho2020terrain, lee2024traversability, kim2025parkour}, showing their potential for various applications. 

Many hydraulic quadruped robots have been operated with model-based control \cite{sirouspour2001ctrl, hyq, minihyq, kuindersma2016atlas, cho2023mpc}. Model-based control is a control strategy that relies on a mathematical model of the system's dynamics. The model predicts the robot's future behavior and optimizes the control inputs accordingly, enabling effective decision-making and real-time adaptation. For example, the hydraulic quadruped robot HyQ\cite{hyq} utilized model-based control to achieve stable locomotion, demonstrating the effectiveness of model-based control in such systems\cite{ugurlu2013dynamic}. Model-based control provides accurate and reliable control using an accurate system dynamics model. However, it can be computationally intensive and less adaptable to dynamic or unpredictable environments.

Recently, the control of quadruped robots has increasingly relied on reinforcement learning (RL) approaches\cite{hwangbo2019learning, tsounis2020quadrupedGait, ji2022concurrent, kim2025parkour}. RL-based control offers greater flexibility, adapting more effectively to changing conditions and learning near-optimal policies without an explicit model, making it more robust in unpredictable scenarios\cite{leeJoonho2020terrain, choi2023deformable}. Despite the advantages of RL-based control, model-based control has been widely used in hydraulic robots\cite{cho2023mpc} due to its ability to manage the complex dynamics of hydraulic systems. The low simulation-to-reality (sim-to-real) transferability in hydraulic systems makes it challenging to deploy RL-based controllers on real-world hardware platforms.

\begin{figure}[!t]
    \centering
    \includegraphics[width=\linewidth]{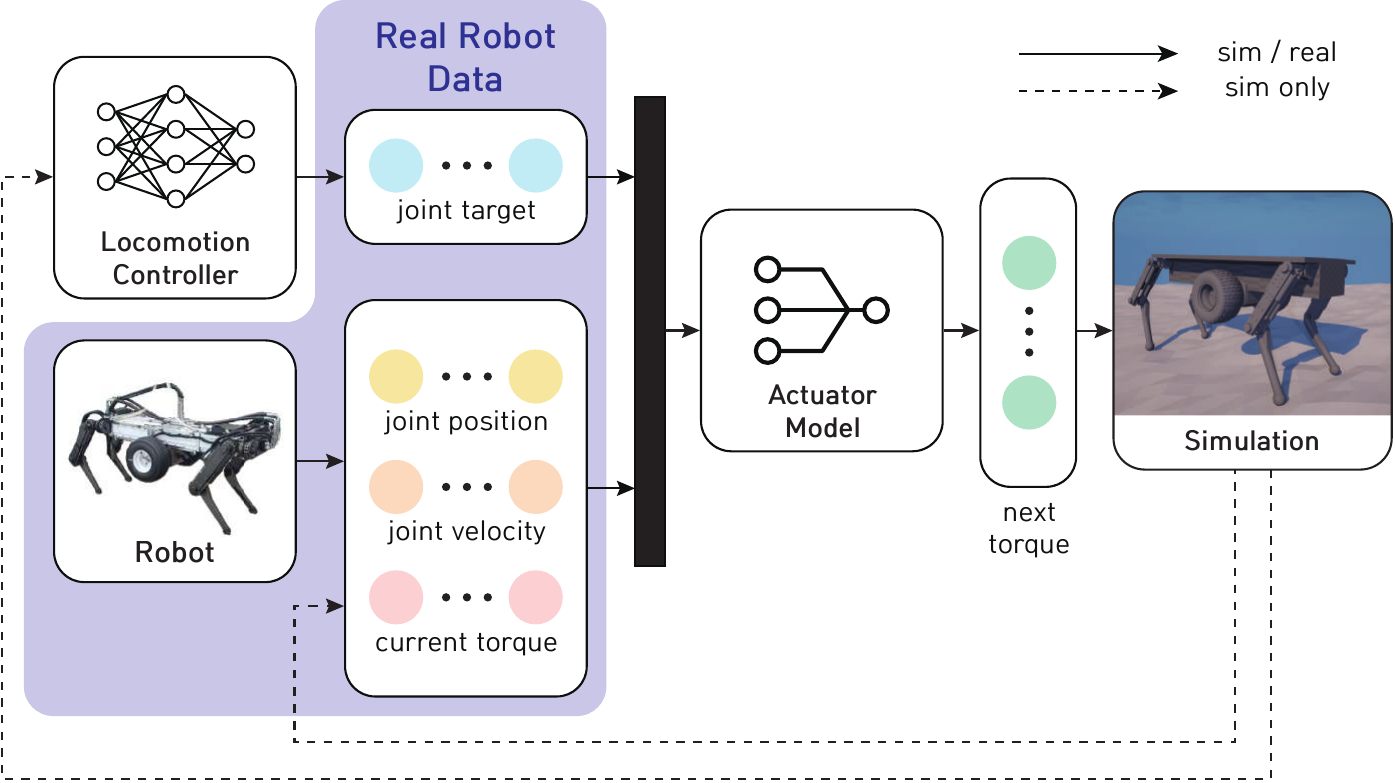}
    \caption{Actuator model framework} 
    \label{fig:actuator_model}
    \vspace{-0.6cm}
\end{figure}

Sim-to-real transfer is considered more difficult in hydraulic systems than in electric motor systems because of the inherent slow response and the complex fluid dynamics of hydraulic systems\cite{sirouspour2001ctrl, hyq, minihyq, cho2023mpc}. Hydraulic robots comprise multiple interconnected cylinders, making accurate simulation difficult due to fluid compressibility and variations in pressure delivery speed in real systems. Furthermore, the large size and heavy weight of the robot can cause undesirable bending and compression of mechanical parts, further complicating the sim-to-real transfer. Moreover, collecting data from such a system without a reliable controller is challenging, as accidents can result in serious damage to the robot.

While previous approaches have informed our work, our method differs in two key ways. First, we propose a simple analytical actuator model derived from hydraulic actuator dynamics, which facilitates sim-to-real transfer. This model accurately captures the behavior of hydraulic systems with negligible computational cost($<1 \mu s$ for all 12 actuators simultaneously). Recent studies often rely on neural network-based actuator models \cite{hwangbo2019learning, kim2021fctrl, wang2024nnctrl}. While these models can provide accurate predictions within the domains they are trained on, they tend to generalize poorly to out-of-distribution scenarios. This poses a serious risk in large-scale hydraulic systems, where untested conditions during data collection can lead to hazardous failures. Moreover, collecting isolated actuator data is impractical for legged robots, as joint loads vary depending on configuration and motion history. These dependencies make it difficult to replicate realistic operating conditions outside the full robotic system. We capture the real-world dynamics of hydraulic actuators—including external loads and fluid effects—by collecting locomotion data directly from a quadruped robot. Building on this, we develop an analytical model capable of handling a broad range of operating conditions to support robust reinforcement learning training. 

Also, we validate our approach on a fully hydraulic quadruped robot. Unlike prior studies that focus on single-actuator hardware demonstrations \cite{guan2008ctrl, kim2021fctrl}, our work showcases the model’s effectiveness in a complete, real-world multi-actuator system. In this study, we make the following key contributions:
\begin{itemize}
    \item Proposal of an analytical actuator model with accurate torque prediction, showing high sim-to-real transferability.
    \item Development of an RL-based locomotion controller for a hydraulic quadrupedal robot weighing over 300kg.
    \item Demonstration of hydraulic-powered RL locomotion at 1m/s, including adaptability to dynamic conditions.
\end{itemize}
To the best of our knowledge, this is the first demonstration of locomotion at 1 m/s on a hydraulic quadruped robot weighing over 300 kg using an RL-based locomotion controller.

\section{RELATED WORKS}

RL-based control approaches have recently gained widespread adoption in the quadruped robot community. Previous work \cite{hwangbo2019learning} demonstrated the successful use of RL for learning agile and dynamic motor skills in legged robots, achieving high performance in complex tasks such as walking, running, and recovering from falls. More recent developments \cite{kumar2021RMA} adopted the RL-trained policies to traverse various terrains, enhancing the sim-to-real transferability and improving robustness in quadruped locomotion. However, it is challenging to apply RL directly to hydraulic quadruped robots due to the difficulty in sim-to-real transferability in hydraulic systems.

Several attempts have been made to improve the sim-to-real transferability between theoretical estimations and actual hydraulic systems\cite{sirouspour2001ctrl, guan2008ctrl, kim2021fctrl,tcheumchoua2022nntorqctrl,cho2023mpc, wang2024nnctrl}. One line of work \cite{guan2008ctrl} proposed a nonlinear adaptive robust control strategy for a hydraulic actuator to estimate system parameters and handle nonlinearities. However, this approach is heavily dependent on the accuracy of the system model and is computationally demanding. Another work \cite{kim2021fctrl} presented a neural network inverse model of a hydraulic actuator to track a force trajectory accurately. However, their method was limited to a single actuator system and was verified only in a disturbance-free environment. Recent advancements \cite{wang2024nnctrl} introduced adaptive online neural predictive control for hydraulic actuators, demonstrating low tracking error and robustness to model uncertainties. However, the use of neural networks requires a large amount of high-quality training data, and real-time application remains challenging due to their high computational cost.

\section{METHODS}

\subsection{Hydraulic quadruped robot}

\begin{figure}[t!]
    \centering
    \includegraphics[width=0.9\linewidth]{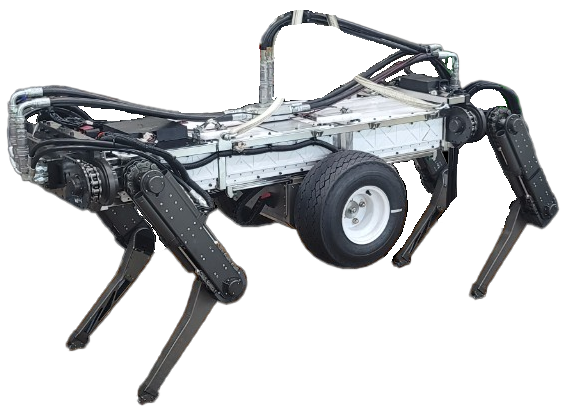}
    \caption{A hybrid-locomotion platform combined with wheeled \& quadruped system}
    \label{fig:robot}
    \vspace{-0.2cm}
\end{figure}

\begin{figure}[t!]
    \centering
    \includegraphics[width=\linewidth]{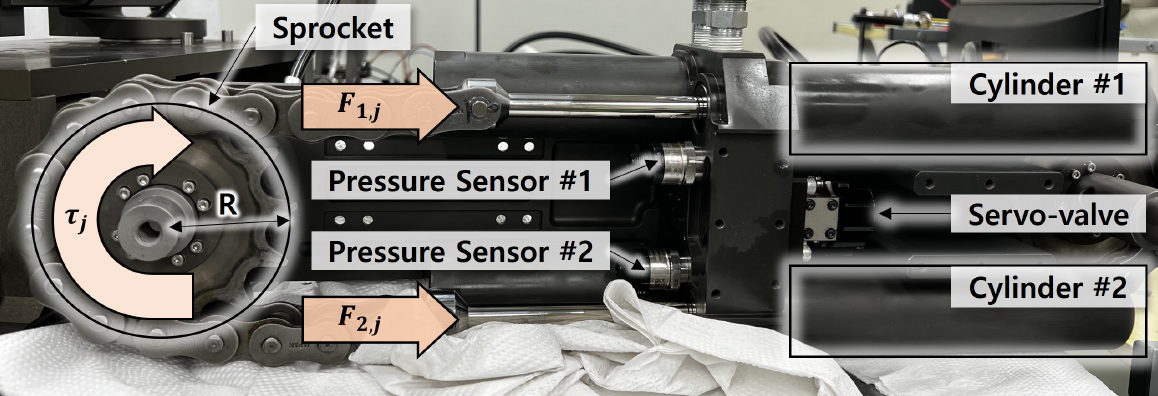}
    \caption{Structure of the hydraulic actuator, where two hydraulic cylinders function as a single unit to generate torque on a revolute joint. The illustration details how chains transfer force from the hydraulic cylinders to the joint. Each leg has three such actuators (roll, pitch, and knee), all following the same structural design.}
    \label{fig:joint}
    \vspace{-0.7cm}
\end{figure}

Our robot is a hybrid locomotion platform used to validate the proposed actuator model and the locomotion performance of a policy trained with the actuator model. Our robot weighs over 300kg and has a length exceeding 1.8 meters. It is a ground vehicle developed to navigate complex terrains while carrying heavy supplies. It features a four-legged design and a two-wheel differential drive system, as shown in Fig. \ref{fig:robot}. The robot can handle various environments, driving at high speeds on flat ground and transitioning to walking when driving is impractical.
The leg mechanism comprises a 3-degree-of-freedom hydraulically actuated joint. Each rotary joint integrates a sprocket and chain mechanism, as shown in Fig. \ref{fig:joint}, which addresses torque nonlinearity issues while having a wide range of motion. Each leg has three revolute actuators—roll, pitch, and knee joint—all of which share the same structural design. The torque for each joint is controlled by adjusting the pressures and displacements within two hydraulic cylinders arranged in a bidirectional pulling structure, which together form a single revolute actuator. A total of 24 hydraulic cylinders control the movement, all of which are connected to a single pump that supplies the fluid from outside the robot. In this study, we focus on the quadrupedal locomotion of the robot. All experiments are conducted with the driving system deactivated.

\subsection{Hydraulic actuator control}

\begin{figure}[t!]
    \centering
    \vspace{0.3cm}
    \includegraphics[width=\linewidth]{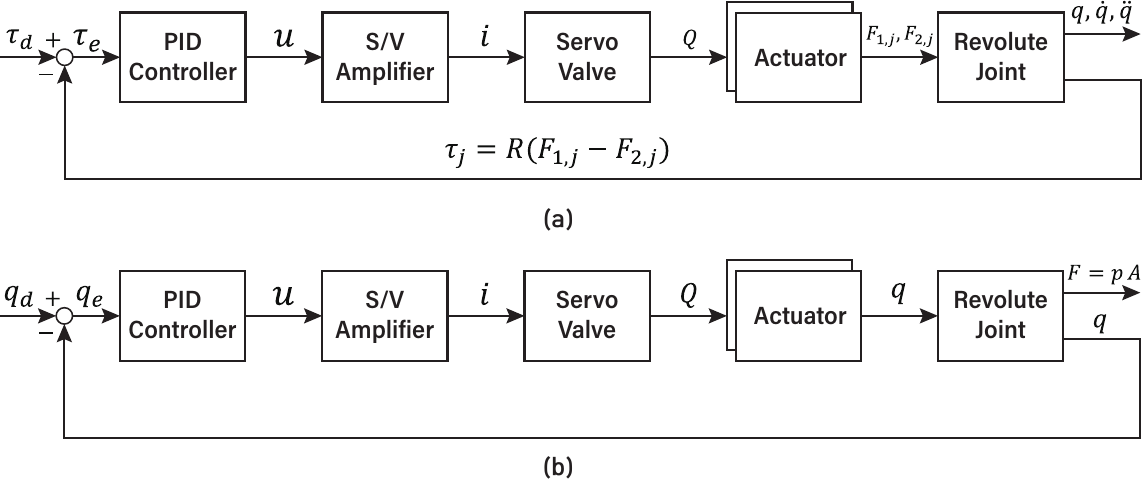}
    \caption{Control diagram of the joint with (a) torque PID control and (b) position PID control}
    \label{fig:act_control}
    \vspace{-0.5cm}
\end{figure}

\begin{figure}[t!]
    \centering
    \includegraphics[width=\linewidth]{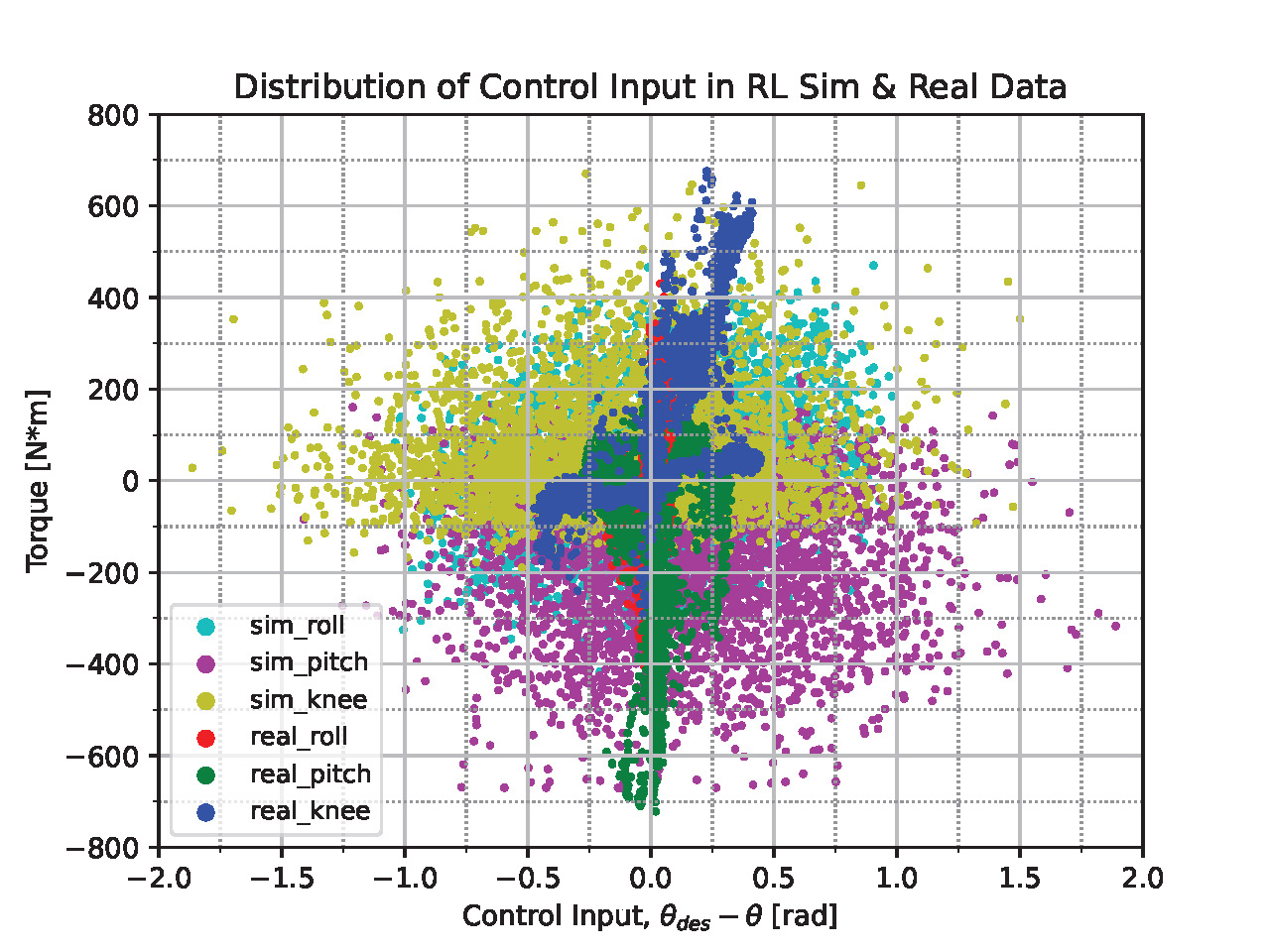}
    \caption{Data distribution of simulation and real data}
    \label{fig:data_distribution}
    \vspace{-0.7cm}
\end{figure}

Each joint operates under two distinct low-level control modes: torque and position PID control, each suited for different tasks. The block diagrams for both controls are shown in Fig. \ref{fig:act_control}, where $u$ is the servo valve input, $i$ is the electric current, $Q$ is the flow rate, and $q$ is the current position of the joint.

Torque PID control adjusts hydraulic pressure to achieve a target torque, making it computationally efficient and suitable for simulations where torque output is assumed to match the target. However, real-world discrepancies arise due to measurement noise and delays, reducing reliability in dynamic tasks like locomotion. Position PID control regulates actuator position for greater stability and precision in real-world applications, as it relies on accurate encoders. While less sensitive to measurement errors, it requires a detailed actuator model for effective simulation, making implementation more complex.

We address these challenges by proposing an actuator model that simplifies hydraulic actuator dynamics, enabling efficient simulation of position control with minimal computational overhead. Our actuator model predicts torque output based on its physical state, allowing rapid and realistic simulation of position control. This solution bridges the gap between simulation and real-world performance, enabling effective RL-based training without complex actuator modeling.

\subsection{Actuator model}

Previous work \cite{kim2021fctrl} proposed a first-order differential equation for a single hydraulic cylinder as

\vspace{-0.5cm}
\begin{gather}
\dot f = g(x) \dot x + h(f,x)x_s  \label{eq:f}\\
\dot x_s = \theta (x_s) + \psi (x_s) u \label{eq:x_s}
\end{gather}
\vspace{-0.5cm}
where,

\vspace{-0.2cm}
\begin{align}
\label{eq:gh}
    g(x) &= -A^2 \beta \left( \frac{1}{V_{0A} + Ax} + \frac{1}{V_{0B} + A(L - x)} \right) \notag \\
    h(f, x) &= C_d w \beta A \sqrt{\frac{P_S - P_T}{\rho} - \text{sgn}(x_s) \frac{f}{\rho A}} \notag \\
    &\quad \cdot \left( \frac{1}{V_{0A} + Ax} + \frac{1}{V_{0B} + A(L - x)} \right).\notag
\end{align}
\vspace{-0.2cm}

Here, $f$ is the force, $x$ is the position of the cylinder, $x_s$ is the valve opening, $u$ is the control input (i.e., the reference position), $\theta(x_s)$, and $\psi(x_s)$ are nonlinear functions related to the valve’s model and controller. The function $\text{sgn}(x_s)$ is the sign function, $L$ is the cylinder stroke, $A$ is the cylinder area, $w$ is the area gradient, $\beta$ is the bulk modulus of the hydraulic fluid, $\rho$ is the oil density, $V_{0A}$ and $V_{0B}$ are the dead volume of each cylinder channel, $C_d$ is the discharge coefficient of the valve, $P_S$ is the supply pressure, and $P_T$ is the return pressure.
By rewriting (\ref{eq:f}) into delta notations, the following equation can be obtained:

\vspace{-0.4cm}
\begin{equation}
\label{eq:delta_f}
    \Delta f = g(x) \Delta x + h(f,x)x_s \Delta t \,.
\end{equation}
\vspace{-0.6cm}

However, (\ref{eq:delta_f}) is formulated for a single cylinder system, which does not account for the dynamic interactions present in a multi-cylinder environment. Direct application of (\ref{eq:delta_f}) to systems with multiple interconnected cylinders, as in our robot, can lead to significant errors due to hydraulic coupling effects. In our case, multiple actuators share a common hydraulic circuit, where fluid compressibility and varying pressure delivery speeds cause discrepancies in the return pressures between the cylinders. Nonlinearity and the number of state variables also require high computational costs, which makes it challenging to apply in real-time simulations with 400 parallel environments. Additionally, in environments where forces are applied in a consistent direction, resistive forces and large impacts could be neglected. In contrast, the direction of force application continuously changes during the locomotion of our robot, making resistive forces and dynamic impacts non-negligible factors that require additional consideration.

\begin{figure*}[!th]
    \centering
    \vspace{0.3cm}
    \includegraphics[width=\textwidth]{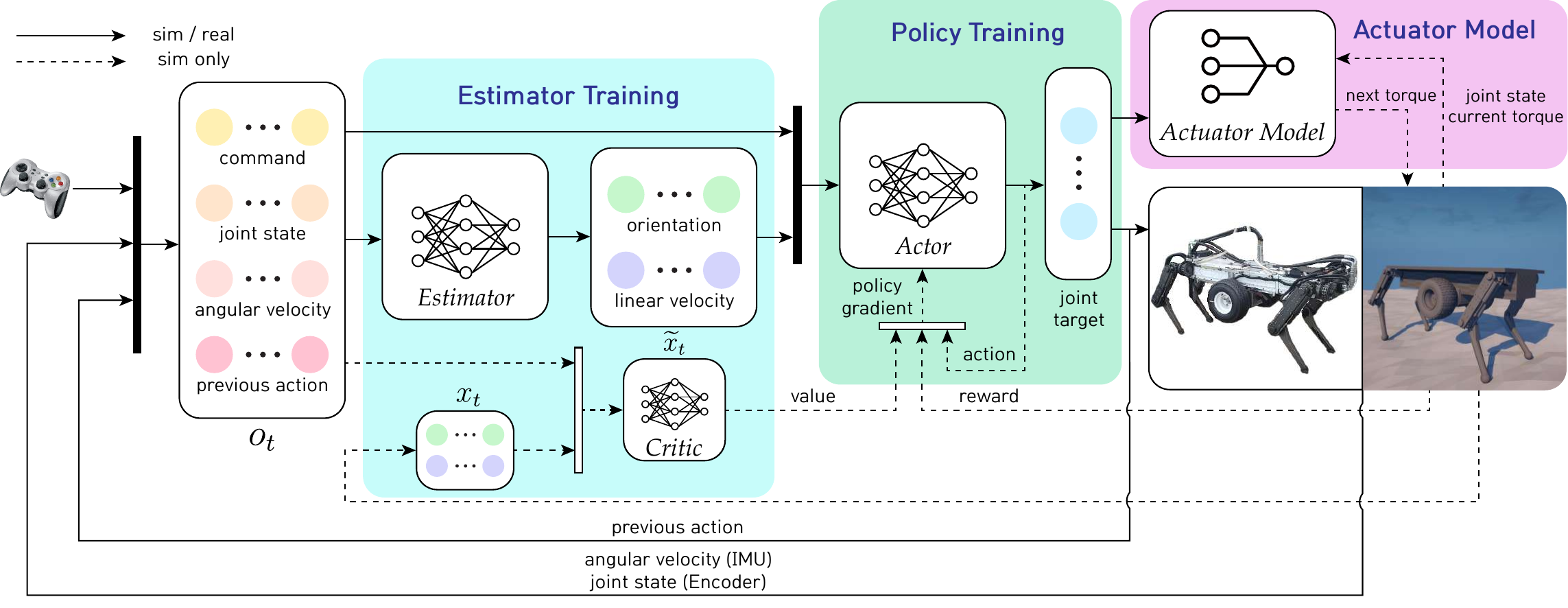}
    \caption{Overall concurrent training framework} 
    \vspace{-0.5cm}
    \label{fig:overall}
\end{figure*}

To address these limitations, we extend the actuator model beyond the single-cylinder formulation (\ref{eq:delta_f}) by introducing additional terms, including $k_3$ to model resistive forces and $k_4$ to account for corrective forces during sudden impacts. Furthermore, we simplify the original formulation while preserving its ability to capture essential hydraulic dynamics, thereby enhancing its practicality for real-world applications. Specifically, we propose the following model:

\vspace{-0.4cm}
\begin{align}
\begin{split}
\label{eq:delta_f_mine}
    \Delta f = k_1&\Delta x - k_2 f - k_3 \dot x \\ & + k_4 \Delta x \cdot \text{max} \left(-f \cdot \text{sgn}(\Delta x), 0\right) \, .
\end{split}
\end{align}
\vspace{-0.4cm}

In (\ref{eq:delta_f}), $g(x)$ represents the position-dependent mechanical leverage effect as a function of $x$. However, its variation remains minimal within the robot's operating range, which can be approximated as a constant $k_1$. \( h(f,x)x_s dt \) represents the force-dependent pressure drop across the hydraulic system. $f$ is scaled 1000 times larger than $x$. The applied force primarily influences this pressure drop, while other factors remain constant or vary negligibly. Thus, it can be approximated as a negative linear function of force, with $k_2$ capturing the system-dependent hydraulic characteristics. This simplification eliminates the need to solve nonlinear differential equations, substantially reducing computational cost while maintaining sufficient accuracy for RL training.

The term \(k_3\dot {x} \) is introduced to model the velocity-dependent resistive force, which was negligible in (\ref{eq:f}) when rapid changes in the direction of force application were minimal. However, in our experiments, the actuator's movement direction changes continuously, making fluid resistance a significant factor. Therefore, incorporating this resistive term is essential to accurately capture the actuator’s real-world behavior.

We introduce the corrective term $k_4$ to account for sudden impact scenarios. The control input is conflicted by the physical dynamics of the actuator when the direction of motion is opposite to the current force. This happens when the actuator is attempting to resist an excessive external load from an impact, yet it is forced to move in a direction opposite to the force. The system's control objective is misaligned with its actual physical response, resulting in discrepancies between the intended control and the resulting motion in the simulation without appropriate correction. Our robot experiences high external loads and reaction forces, and these effects are not adequately accounted for in the previous model (\ref{eq:f}). To address this issue, we introduce a corrective force that adjusts the control input in these scenarios. This term prevents excessive reductions in control force, ensuring that the model’s responses remain consistent with the system’s actual dynamics, even in the presence of large external disturbances or impacts.

The derivation of the corrective term begins with a reconsideration of the fundamental hydraulic pressure dynamics governing the actuator. The actuator force $f$ can be represented as $f=p\cdot A$ with the pressure $p$ and the effective actuator area $A$. The hydraulic pressure dynamics within the actuator chamber, subject to fluid compressibility and instantaneous displacement demands, is given by:

\vspace{-0.3cm}
\begin{equation}
\frac{dp}{dt} = \frac{B}{V}\left(Q_{in} - A \dot{x}\right).
\label{eq:pressure_dynamics}
\end{equation}
\vspace{-0.4cm}

$B$ denotes the fluid bulk modulus, $V$ is the actuator chamber volume, $Q_{in}$ is the volumetric inflow. In sudden impact scenarios, a rapid displacement command $\Delta x$ leads to an instantaneous and significant inflow demand, described by $Q_{in}\approx A\Delta x / \Delta t$.

We hypothesize that under impact conditions, the commanded displacement rate $\Delta x / \Delta t$ can be considered significantly larger than the actual actuator velocity $\dot{x}$. The actuator piston motion is constrained by flow capacity and oil viscosity, which limit its acceleration under sudden commands. Consequently, while the commanded velocity reflects the instantaneous displacement demand, the actual piston velocity remains much smaller at the initial moment of impact. Hence, the second term can be regarded as negligible, leading to the approximation $\Delta p \approx (BA/V) \Delta x$.

Furthermore, we assume that the pressure variation caused by an impact should scale with the applied impact force. To capture this effect, we introduce a dimensionless normalization with respect to the maximum actuator force, thereby hypothesizing that the corrective term can represent the nonlinear and transient nature of hydraulic responses under sudden impact conditions. The resulting equation is expressed as:

\vspace{-0.5cm}
\begin{equation}
\Delta f_{\text{imp}} = \frac{B A^2}{V f_\text{max}} f \Delta x.
\end{equation}
\vspace{-0.4cm}

The final adjusted term, designed to activate under actual impact conditions (i.e., when \( f\cdot\Delta x < 0 \)), is given by $k_4 \Delta x \cdot \text{max}\left(-f \cdot \text{sgn}(\Delta x), 0\right)$, as shown in (\ref{eq:delta_f_mine}). This corrective force plays a critical role in obtaining a stable RL policy. During RL exploration in simulation, the robot frequently encounters situations involving sudden impacts. The RL actor undergoes various attempts and experiences numerous falls throughout the process of learning locomotion policies, resulting in frequent and substantial impacts. This is shown in Fig. \ref{fig:data_distribution}, where simulation data exhibits a broader distribution compared to real data, particularly when the torque and control input are in opposite directions. If the actuator model is constructed without properly accounting for impacts, it will predict entirely incorrect values when encountering these situations, thereby hindering the RL training process. Additionally, our analytical model can estimate $k_4$ even with a small amount of data, whereas learning-based methods rely on large and diverse input data. Since curve fitting requires significantly fewer data points than neural network training, our model remains reliable despite lower occurrences of impact cases.

By applying $\Delta f = f_{\text{next}} - f$ and the relationships $\tau = R f$, $\Delta x = x_\text{des} - x$, $\dot{x} = R \dot{q}$, and $x = Rq$, where $R$ is the radius of the sprocket, the final expression of the actuator model is obtained from (\ref{eq:delta_f_mine}) as:

\vspace{-0.5cm}
\begin{align}
\label{eq:torq}
    \tau_{\text{next}} =& k_1 R^2  (q_\text{des}-q) + (1-k_2) \tau - k_3 R^2 \dot q \notag \\ & + k_4 R (q_\text{des}-q) \cdot \text{max} \left(-\tau \cdot \text{sgn} (q_\text{des}-q), 0\right) .
\end{align}
\vspace{-0.4cm}

Therefore, (\ref{eq:torq}) shows that our model can predict $\tau_\text{next}$ corresponding to the current torque $\tau$, current joint state $q$ and $\dot q$, and the target position $q_\text{des}$.

The coefficients of the actuator model from (\ref{eq:torq}) are obtained by fitting the data from real-world robot operations. We collected data from operating the robot for 20 seconds with a locomotion policy trained without an actuator model. The locomotion policy was operated at a frequency of 100Hz, while joint positions, target positions, joint velocities and torques were recorded at a higher frequency of 1000Hz. The high-frequency data collection allowed us to capture detailed and precise information about the actuator dynamics in real-robot operations.

We implemented three baseline neural network architectures—MLP, LSTM, and GRU—under the same dataset and training conditions as the proposed actuator model to ensure a fair comparison. Each network was configured with size (48, 64, 12), and hyperparameters such as hidden layer size were tuned to achieve the best accuracy. Training was performed for 1,000 iterations until convergence, providing sufficient optimization.

\subsection{Estimator and policy training with actuator model}

\begin{table}[t!]
\vspace{0.3cm}
    \caption{Global Rewards}
    \vspace{-0.3cm}
    \label{table:global_rew}
    \centering
    \setlength{\tabcolsep}{4pt} 
    \renewcommand{\arraystretch}{1.5} 
    \resizebox{\linewidth}{!}{%
    \begin{tabular}{|c|l|}
        \hline
        \textbf{Reward} & \multicolumn{1}{c|}{\textbf{Expression}}  \\
        \hline
            Command & 
                $\begin{aligned}
            r_v = k_\text{cmd} &\{\exp(-\|\text{cmd} _{xy} - v_{xy}\|^2) \\ 
                &\cdot (1+\exp(-0.5||\text{cmd} _{xy}-v_{xy}||)) \\
                & +\exp(-1.5(\text{cmd}_z - \omega_z)^2)\}
        \end{aligned}$ \\
            \hline
            Yaw Command Error & 
                $\begin{aligned}
            r_\text{yaw}  = 
                \begin{cases} 
                    10c_f k_\text{yaw} (\text{cmd}_z - \omega_z)^2 , & \text{if } \omega_z=0, \\
                    c_f k_\text{yaw} (\text{cmd}_z - \omega_z)^2 , & \text{else}.
                \end{cases}
            \end{aligned}$ \\
            \hline
            Base Height & $r_h = k_h \exp(-40|h_0-h|)$ \\
            \hline 
            Base Motion & $r_m = k_m (v_z^2+0.02(|\omega_x|+|\omega_y|))$ \\ 
            \hline 
            Torque & $r_\tau = c_f k_\tau ||\tau||^2$ \\
            \hline
            Torque Clip & $r_{\tau_\text{clip}}=c_f k_{\tau_\text{clip}} ||\tau_\text{clip}||^2$ \\
            \hline
            Nominal Position & 
                $\begin{aligned}
                r_q  = 
                    \begin{cases} 
                          10c_f k_q ||q_t - q_\text{nom}||, & \text{if }|\text{cmd}|=0, \\
                        c_f k_q ||q_t - q_\text{nom}||, & \text{else}.
                    \end{cases}
                \end{aligned}$ \\
            \hline
            Joint Velocity & $r_{\dot q} = c_f k_{\dot q} ||\dot q_t||^2$ \\
            \hline
            Joint Acceleration & $r_{\ddot q} = c_f k_{\ddot q} ||{\dot q_t}-{\dot q_{t-1}}||^2$ \\
            \hline
            Action smoothness & $\begin{aligned}
        r_{s} = c_f k_{s} (0.5||q_t^{des} - & 2q_{t-1}^{des} + q_{t-2}^{des}||^2 \\ & 
            + ||q_t^{des}-q_{t-1}^{des}||^2)
        \end{aligned}$ \\
            \hline
        \end{tabular}%
    }
    \vspace{-0.2cm}
\end{table}

\begin{table}[t!]
    \caption{Local Rewards}
    \vspace{-0.3cm}
    \label{table:local_rew}
    \centering
    \setlength{\tabcolsep}{4pt} 
    \renewcommand{\arraystretch}{1.5} 
    \resizebox{\linewidth}{!}{%
        \begin{tabular}{|c|l|}
            \hline
            \textbf{Reward} & \multicolumn{1}{c|}{\textbf{Expression}}  \\
            \hline
            Flight Phase & $r_\text{fl} = c_f k_\text{fl} \: (\text{if } \forall i,  \text{contact}_i =\text{False})$ \\
            \hline
            \multicolumn{1}{|c }{Airtime } & \\
            \hline
            \multicolumn{2}{|c|}{
                $\begin{aligned}
                r_\text{air, i} = 
                \begin{cases} 
                    k_a \min(\max(T_{s, i} - T_{a, i}, -0.25),0.25) & \text{if }||\text{cmd}||=0, \\
                    k_a \min(T_{a, i}, 0.2) & \text{if } T_{a, i} < 0.25, \\
                    k_a \min(T_{s, i}, 0.2) & \text{if } T_{s, i} < 0.25, \\
                    0 & \text{else .} \\
                \end{cases}
                \end{aligned}$
            } \\
            \hline
            Foot Slip & $\begin{aligned}
            r_{\text{slip},i} = c_f k_\text{slip} ||v_{\text{foot},i, xy}||^2 \: (\text{if } \text{contact}_i =\text{True}) \\
            \end{aligned}$ \\
            \hline
            Foot Clearance 1 & $\begin{aligned}
            r_{\text{c1}, i} = k_\text{c1} &||v_{\text{foot},i}||(h_i - h_\text{tar})^2 \\
            & (\text{if } \text{contact}_i =\text{False} \: \& \:||\text{cmd}||>0)            \end{aligned}$  \\
            \hline
            Foot Clearance 2 & $r_{\text{c2}, i} = k_\text{c2} r_{\text{c1},t-1,i}  \quad (\text{if } \text{contact}_i =\text{True}) $ \\
        \hline
            GRF Smoothness & $\begin{aligned}
        r_\text{grf, i} = c_f k_\text{grf} \{ 0.5(f_{t,i}^{grf}&-2f_{t-1,i}^{grf} +f_{t-2,i}^{grf})^2 \\ & + (f_{t,i}^{grf} - f_{t-1,i}^{grf})^2\}
        \end{aligned}$ \\
            \hline
            Action Clip & $r_\text{act, j} = c_f k_\text{act} |q_{t,j}^{des, \text{clip}}|$ \\
            \hline
            Joint Limit & $r_{l, j} = c_f k_l \: (\text{if } q_{t,j} >\text{limit})$ \\
            \hline
        \end{tabular}%
    }
    \vspace{-0.6cm}
\end{table}

The locomotion policy of the quadruped robot is trained with RL in 400 parallel environments with flat terrain. The initial state is assigned randomly for each environment, and curriculum learning is employed with factor $c_f$ to facilitate smooth state transitions. Concurrent training\cite{ji2022concurrent} is introduced to train a robust locomotion policy and the state estimator for our robot, as the overall training framework is shown in Fig. \ref{fig:overall}.

Reward functions and coefficients are applied as shown in Table \ref{table:global_rew} and \ref{table:local_rew}. Global rewards are calculated for the whole robot, and local rewards are calculated for each joint or foot. The total reward is calculated by summing all the global and local rewards. The simulation terminates during training when the robot makes contact with the ground, except for its feet and shanks. Observation and kinematics randomization are applied to be robust against disturbances from real-world experiments.

\begin{table*}[!t]
\vspace{0.3cm}
\caption{RMSE and MAPE Loss for Actuator model and MLP, LSTM, GRU}
\label{table:actmodel_loss}
\renewcommand{\arraystretch}{1.3}
\resizebox{\textwidth}{!}{
    \begin{tabular}{|c|c|c|c|c|c|c|c|c|c|c|c|c|c|c|c|}
        \hline
        \multirow{3}{*}{Models} & 
        \multicolumn{3}{c|}{Command 0.4 m/s} &
        \multicolumn{3}{c|}{Command  1.0 m/s (1)} &
        \multicolumn{3}{c|}{Command  1.0 m/s (2)} &
        \multicolumn{3}{c|}{Command  1.0 m/s (3)} &
        \multicolumn{3}{c|}{Command  1.0 m/s (4)} \\
        \cline{2-16}
        &
        \multirow{2}{*}{RMSE}  &
        \multicolumn{2}{c|}{$|\tau_j|>50$} &
        \multirow{2}{*}{RMSE}  &
        \multicolumn{2}{c|}{$|\tau_j|>50$} &
        \multirow{2}{*}{RMSE}  &
        \multicolumn{2}{c|}{$|\tau_j|>50$} &
        \multirow{2}{*}{RMSE}  &
        \multicolumn{2}{c|}{$|\tau_j|>50$} &
        \multirow{2}{*}{RMSE}  &
        \multicolumn{2}{c|}{$|\tau_j|>50$} \\
        \cline{3-4} \cline{6-7} \cline{9-10} \cline{12-13} \cline{15-16}
        & 
        &  
        RMSE & MAPE [\%] &
        &  
        RMSE & MAPE [\%]&
        &  
        RMSE & MAPE [\%]&
        &  
        RMSE & MAPE [\%]&
        &  
        RMSE & MAPE [\%]\\
        \hline \hline
            
        \textbf{Actutator model (Ours)} & 
        \textbf{7.45} & \textbf{8.06} & \textbf{4.22} &
        \textbf{8.44} & \textbf{9.15} & \textbf{4.77} &
        \textbf{7.90} & \textbf{8.67} & \textbf{4.48} &
        \textbf{10.6} & \textbf{11.4} & \textbf{5.65} &
        \textbf{10.8} & \textbf{11.7} & \textbf{5.72} \\
        \hline
        MLP & 
        36.0 & 41.7 & 20.9 &
        47.1 & 53.5 & 27.3 &
        38.9 & 44.8 & 22.8 &
        57.5 & 64.4 & 33.2 &
        57.7 & 63.8 & 32.6 \\
        \hline
        LSTM & 
        33.3 & 37.2 & 19.8 &
        41.1 & 46.5 & 23.8 &
        35.3 & 39.7 & 20.6 &
        50.1 & 55.8 & 28.2 &
        51.2 & 56.9 & 28.7 \\
        \hline
        GRU & 
        27.4 & 30.7 & 15.6 &
        34.4 & 38.3 & 19.9 &
        28.9 & 32.2 & 16.4 &
        41.8 & 46.1 & 24.0 &
        42.1 & 45.9 & 24.0 \\
        \hline
    \end{tabular}
}
\vspace{-0.5cm}
\end{table*}

The RL training network consists of three parts: actor, critic, and estimator. All three networks are designed using MLP. The estimator estimates the current robot state $x_t$ from the observation $o_t$, and the actor receives the estimated robot state $\widetilde{x}_t$ with $o_t$ and outputs an action $a_t$, which is the desired joint position. The observation state $o_t$ consists of 
\begin{align}
\label{eq:observation}
    o_t = (\omega, q_t, \dot q_t, q_{t-1}^{des}, \text{cmd} )
\end{align}
where $\omega$ stands for the base angular velocity, $q_t$ and $\dot q_t$ are the joint states of time t, $q_{t-1}^{des}$ is the previous action which is the desired joint position from the previous time step, and $\text{cmd}$ is the velocity command. The critic outputs a value from $o_t$ and $x_t$, and the policy is trained with PPO. The estimator is trained by supervised learning with $x_t$ and $\widetilde{x}_t$. The locomotion policy and the estimator are operated at 100Hz simultaneously, while the simulation and the position controller in the real robot are operated at 1000Hz. The training is conducted in Raisim\cite{raisim} using an AMD Ryzen Threadripper 3990X and NVidia RTX 3090, with 20,000 iterations taking around 10 hours.

\section{RESULTS}

\begin{figure}[t!]
    \centering
    \includegraphics[width=\linewidth]{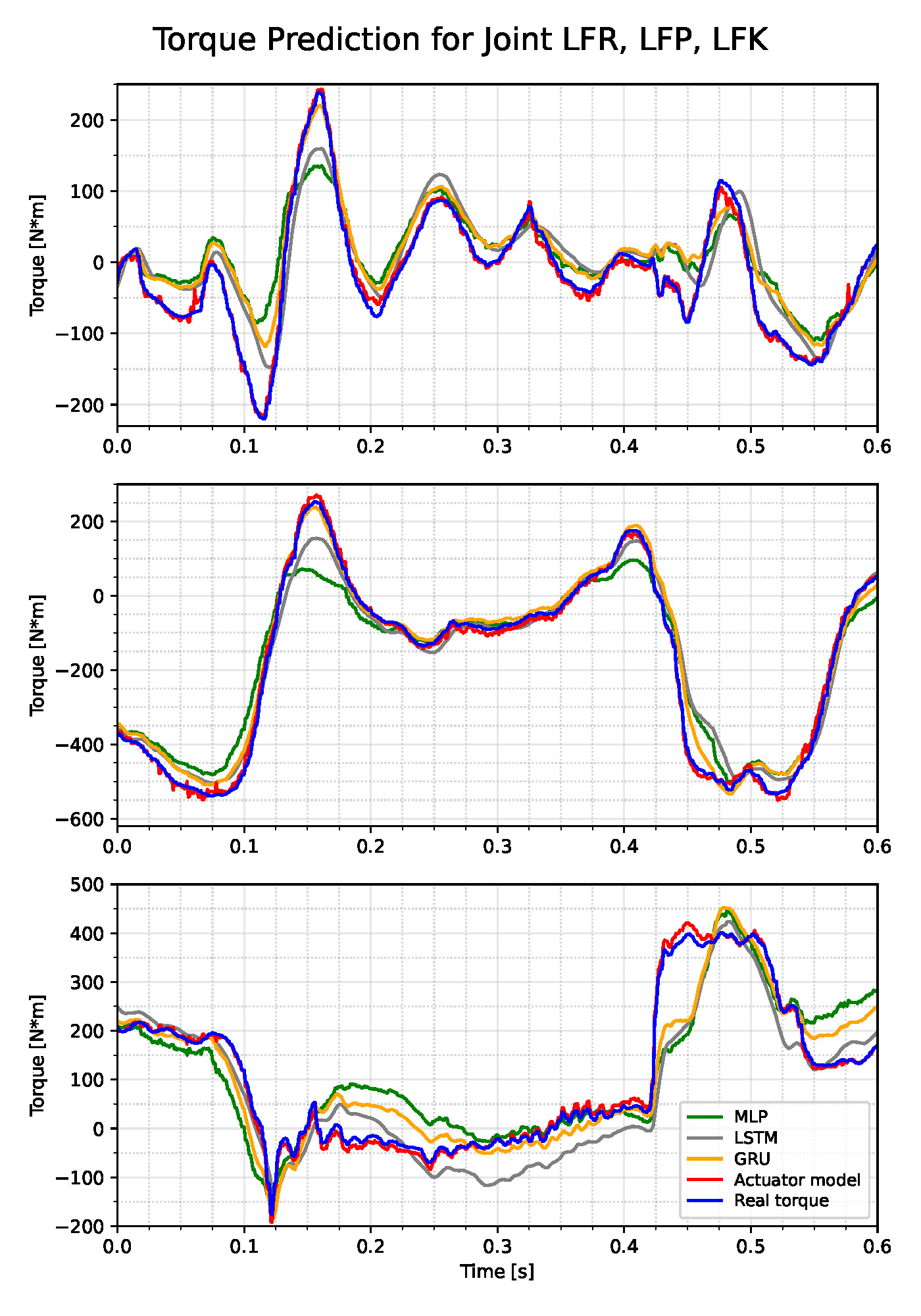}
    \caption{Torque prediction of the actuator model of joint LFR, LFP, LFK}
    \label{fig:actnet}
    \vspace{-0.4cm}
\end{figure}

\begin{figure}[t!]
    \centering
    \includegraphics[width=\linewidth]{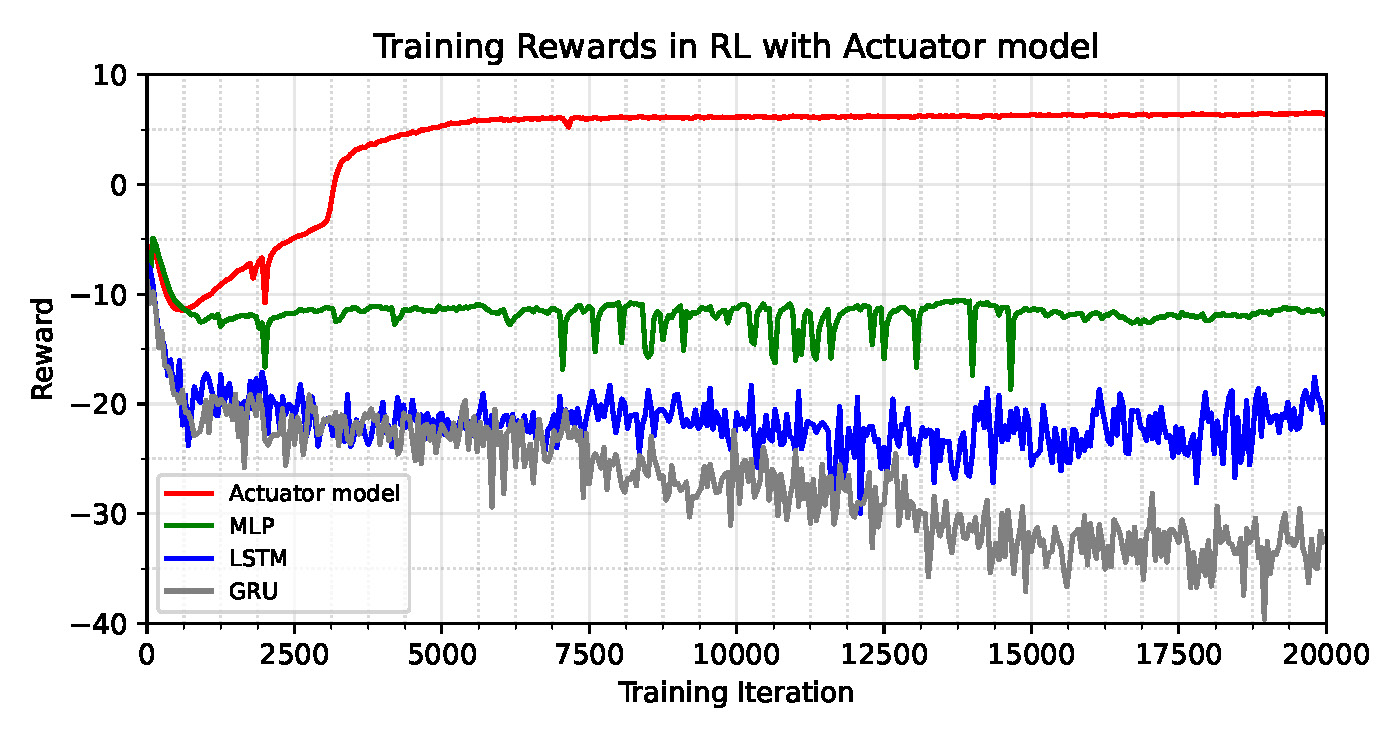}
    \caption{Reward shown according to the training iteration.}
    \label{fig:train_res}
    \vspace{-0.6cm}
\end{figure}

\subsection{Evaluation of actuator model}

We compared the proposed actuator model’s prediction results with the baseline neural networks (MLP, LSTM, and GRU) under the conditions described in Section III-C. In total, five real-robot locomotion experiments were conducted. One trial was performed with a forward command of 0.4 m/s, while the remaining four were conducted at a forward command of 1.0 m/s. During these experiments, the locomotion policy generated target positions at 100Hz, and $o_t$ were recorded at 1000Hz. We compared the predicted torques with the measured torques at each timestep using the collected data. RMS and Mean Absolute Percentage (MAP) errors with a threshold of 50 N$\cdot$m for the actual torque values were calculated. This threshold is introduced because the error becomes disproportionately more prominent in the MAPE metric when the actual torque values are small. Table \ref{table:actmodel_loss} summarizes the results for RMS and MAP errors, whereas Fig. \ref{fig:actnet} also shows that our model has the highest accuracy in predicting the torque values. Our model outperforms the trained neural networks in terms of torque prediction accuracy across all experimental data, demonstrating its performance in sim-to-real transfer. This accuracy enables the RL-based controller to operate reliably across a broader range of conditions, ensuring stable and robust locomotion. Unlike neural networks, which struggle with out-of-distribution data, our actuator model provides consistent torque predictions even in dynamic environments, allowing the RL policy to adapt to disturbances and variations in terrain.

\begin{figure}[!t]
    \vspace{0.3cm}
    \centering
    \includegraphics[width=\linewidth]{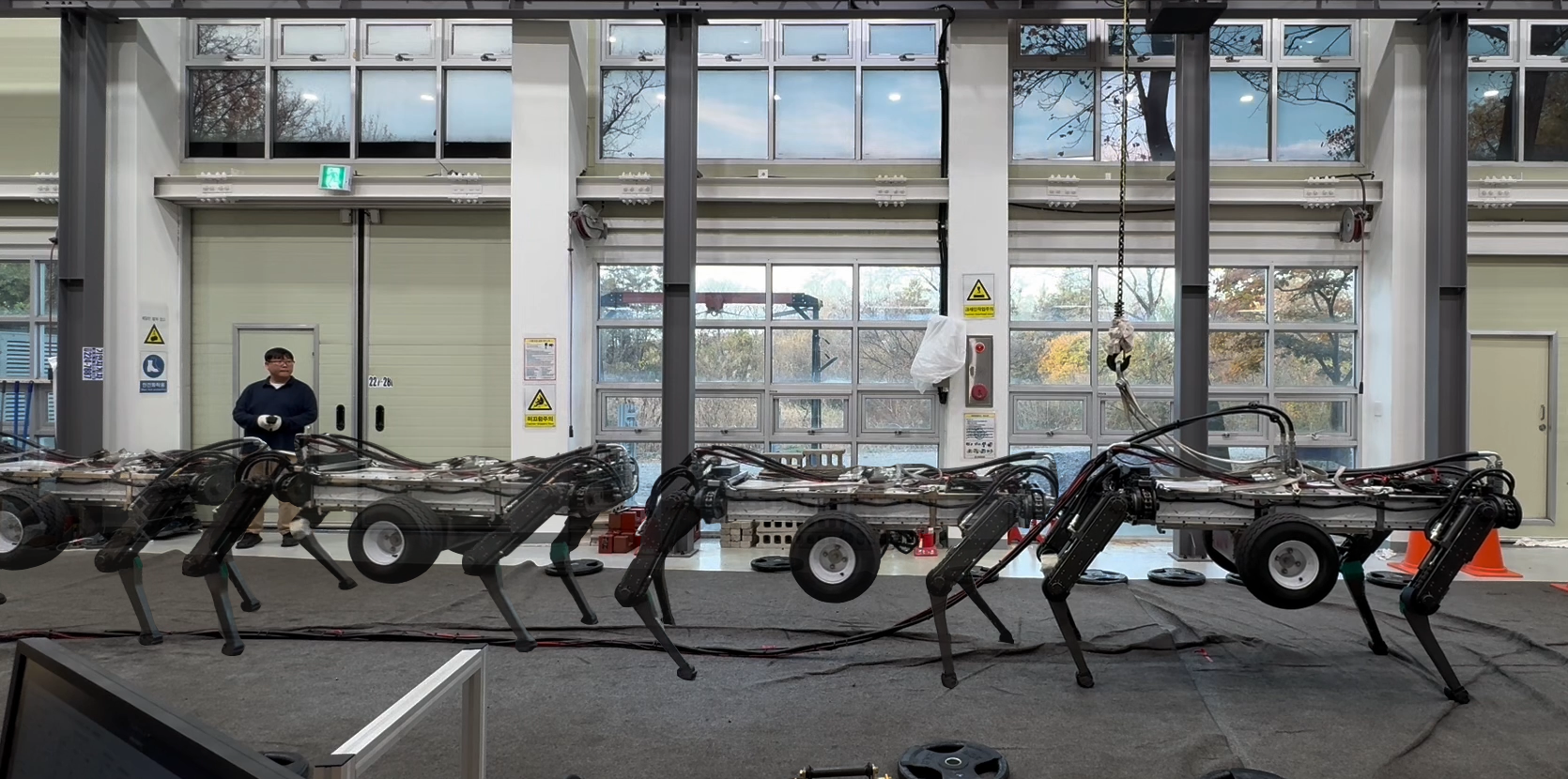}
    \caption{Locomotion results with 1.0 m/s command, trained with our actuator model.}
    \label{fig:locomotion_res_1.0}
    \vspace{-0.4cm}
\end{figure}

\begin{figure}[t!]
    \centering
    \includegraphics[width=\linewidth]{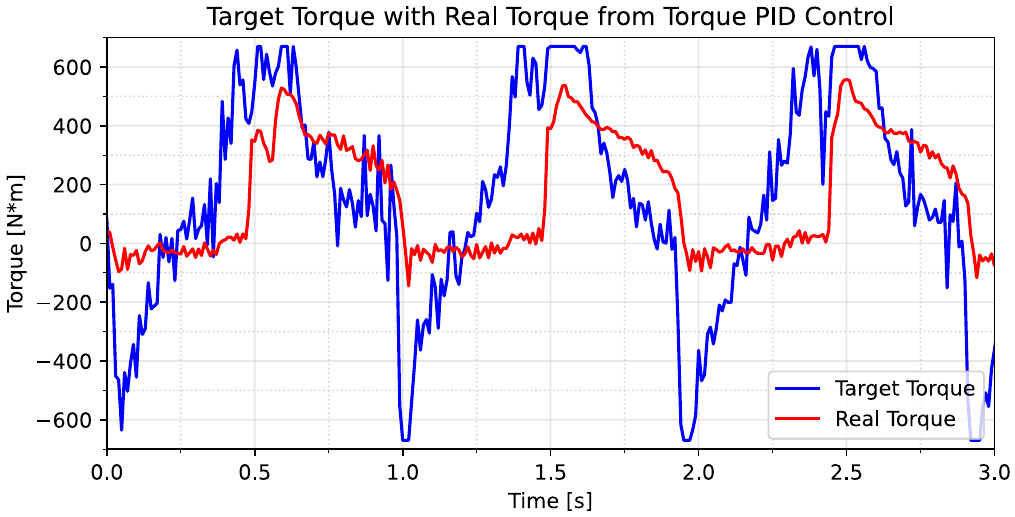}
    \caption{Torque control results, actual vs target torque with torque control}
    \label{fig:torq_control_res}
    \vspace{-0.7cm}
\end{figure}

In addition, we implemented the proposed actuator model and baseline models(MLP, LSTM, GRU) in the simulation and trained locomotion policies with RL. The reward values recorded during training are shown in Fig. \ref{fig:train_res}. Notably, our actuator model outperformed the exploration process in RL, as the latter failed to achieve appropriate reward values and did not converge to a stable result.

The proposed model's performance arises from its ability to generalize beyond the narrow operational range of the training data. Although the dataset for the actuator model was limited, our model successfully predicted the corresponding torques in unstable situations. The baseline models were also trained with the same data but faced difficulties, often failing to generate reasonable outputs in unstable situations and consequently breaking down during RL training. To learn dynamic locomotion successfully, the actuator model must handle exceptional cases such as falls, where large impact forces frequently occur. Unlike the neural networks trained solely on standard walking data, the proposed actuator model incorporates the $k_4$ term, which applies corrective forces under impact conditions based on the actuator's physical state. This mechanism enabled the model to remain accurate in out-of-distribution scenarios and effectively respond to dynamic disturbances, thereby ensuring robustness during training and real-world applications. Our model's performance demonstrated its ability to generalize beyond the training data and handle complex, non-ideal situations, which is critical for robust quadruped robot locomotion.

We also compared the computational cost of training. The proposed actuator model required 10 hours to train the locomotion policy with RL, whereas the baseline models required between 36 and 48 hours under the same training conditions. This result highlights that our model achieves superior accuracy while being more than three times faster to train, thereby offering advantages from both performance and efficiency perspectives.

\subsection{Evaluation of the locomotion performance}

We compared the robot's locomotion performance with the trained policies using three different control approaches. These included torque PID control trained in RL without an actuator model, position PID control trained in RL with stiff PD estimation instead of an actuator model, and position PID control trained in RL with the proposed actuator model. All experiments were conducted indoors on a flat floor, where the robot walked about 30 meters in a straight trajectory to provide a controlled evaluation environment for the trained policies. The robot's walking performance was evaluated in two main areas: command tracking and locomotion stability. The supplementary material includes detailed experimental videos that illustrate these approaches.

\subsubsection{Command Tracking}

The robot was tested with tracking commands of 0.4 m/s and 1.0 m/s. The torque control demonstrated satisfactory performance at a speed of 0.4 m/s, with the command being tracked effectively. However, as the command speed increased, torque control struggled to keep the torque output in sync with the target torque, as shown in Fig.  \ref{fig:torq_control_res}, causing the robot to lose balance and tip forward. The robot struggled to follow the target torque in torque control, resulting in failures during dynamic locomotion. In contrast, the position controls followed the commands up to 1.0 m/s, demonstrating improved robustness to higher speeds. Notably, the locomotion controller trained with the actuator model showed sufficient capability, as in Fig. \ref{fig:locomotion_res_1.0}, to increase the command speed beyond 1.0 m/s. This highlights the advantage of using an accurate actuator model, which not only enhances locomotion stability but also enables smoother control transitions, even in untrained conditions. However, this could not be thoroughly tested due to spatial constraints in the experimental setup and limitations related to the crane's speed, which was used to prevent the robot from tipping over. Nevertheless, it is expected that the locomotion controller trained with the actuator model would have been able to handle even higher command speeds under more favorable conditions.

\subsubsection{Locomotion stability}

The torque control exhibited significant issues with locomotion stability as the command speed increased, resulting in the robot failing to maintain stability. The actuator delays shown in Fig. \ref{fig:torq_control_res} caused the robot to walk with pronounced impacts, resulting in noticeable vibrations and sounds. The robot frequently lost stability and fell when its feet encountered obstacles or the ground. The position control without the actuator model showed stability during locomotion but faced challenges due to its stiff nature of high-gain PD estimation. This stiffness restricted the control's ability to lift the legs sufficiently, especially when encountering obstacles such as uneven surfaces. This limitation resulted in instability and compromised locomotion performance in more complex environments, as holding the foot slightly above the ground. Moreover, the stiff nature of the controller resulted in a less natural walking pattern, with sudden vibrations to the robot's body. However, the position control trained with the proposed actuator model demonstrated superior locomotion stability. The locomotion was natural and smooth, with the robot's feet reaching sufficient height for stable movements, as exactly shown inside the simulation environment. The learned policy demonstrated the ability to perform turning and lateral walking on flat terrain without requiring retraining, further supporting its robustness under the current training conditions. Overall locomotion was stable, and the robot exhibited more fluid and adaptive movements, even in dynamic conditions with sudden disturbances. This adaptivity was facilitated by the RL-based controller, which leveraged the accurate torque estimation of the actuator model to develop more robust control policies.

These results demonstrate that the proposed actuator model not only accurately simulates the behavior of hydraulic actuators but also enables the seamless transfer of simulation-based policies to real-world robotic systems. The stable and robust locomotion confirms the effectiveness of our actuator model under a wide range of conditions, both in simulation and real-world experiments.

\section{CONCLUSION}

This study demonstrated the effectiveness of the proposed actuator model by showing both accurate torque prediction and improvement of real-world locomotion. Our actuator model enabled the development of a robust RL-locomotion controller, demonstrating successful sim-to-real transfer in heavy hydraulic robots. The RL policy trained with the actuator model outperformed traditional methods, delivering smooth and stable locomotion both in simulation and on the real robot, even at higher speeds and under dynamic conditions.

However, the model's performances were not tested in scenarios that deviate significantly from the training data, such as abrupt directional changes or extreme external forces. Additionally, the long-term durability of the actuator model remains unassessed, and the experimental setup limited testing beyond 1.0 m/s.

Future work will focus on expanding real-world testing to include a broader range of conditions, such as higher speeds and diverse terrains, and refining the actuator model by incorporating factors such as actuator wear and hydraulic variations to ensure long-term reliability. We also aim to evaluate the model's adaptability to different robotic platforms and tasks, increasing its generalizability beyond the quadruped robot used in this study. In the long term, we aim to develop controllers that combine the robustness of RL with the interpretability of model-based approaches, contributing to more reliable deployment in real-world systems.

\addtolength{\textheight}{-1cm}   





\bibliographystyle{IEEEtran}
\bibliography{ref}

\end{document}